\documentclass[conference]{IEEEtran}
\IEEEoverridecommandlockouts
\usepackage{cite}
\usepackage{amsmath,amssymb,amsfonts}
\usepackage{algorithmic}
\usepackage{algorithm}
\usepackage{graphicx}
\usepackage{textcomp}
\usepackage{tabularx}
\usepackage{xcolor}
\usepackage{booktabs}
\usepackage{array,ragged2e}

\pdfoutput=1

\def\BibTeX{{\rm B\kern-.05em{\sc i\kern-.025em b}\kern-.08em
    T\kern-.1667em\lower.7ex\hbox{E}\kern-.125emX}}
\begin{document}

\title{HDF: Hybrid Deep Features for Scene\\ Image Representation
}

\author{\IEEEauthorblockN{ Chiranjibi Sitaula}
\IEEEauthorblockA{\textit{School of IT} \\
\textit{Deakin University}\\
Geelong, Australia \\
csitaul@deakin.edu.au}
\and
\IEEEauthorblockN{Yong Xiang}
\IEEEauthorblockA{\textit{School of IT} \\
\textit{Deakin University}\\
Geelong, Australia \\
yong.xiang@deakin.edu.au}
\and
\IEEEauthorblockN{Anish Basnet}
\IEEEauthorblockA{\textit{Department of IT} \\
\textit{Ambition College}\\
Kathmandu, Nepal \\
anishbasnetworld@gmail.com}
\and
\IEEEauthorblockN{Sunil Aryal}
\IEEEauthorblockA{\textit{School of IT} \\
\textit{Deakin University}\\
Geelong, Australia \\
sunil.aryal@deakin.edu.au}
\and
\IEEEauthorblockN{Xuequan Lu}
\IEEEauthorblockA{\textit{School of IT} \\
\textit{Deakin University}\\
Geelong, Australia \\
xuequan.lu@deakin.edu.au}

}

\maketitle

\begin{abstract}
Nowadays it is prevalent to take features extracted from pre-trained deep learning models as image representations which have achieved promising classification performance. Existing methods usually consider either object-based features or scene-based features only. However, both types of features are important for complex images like scene images, as they can complement each other. 
In this paper, we propose a novel type of features -- hybrid deep features, for scene images.
Specifically, we exploit both object-based and scene-based features at two levels: part image level (i.e., parts of an image) and whole image level (i.e., a whole image), which produces a total number of four types of deep features. Regarding the part image level, we also propose two new slicing techniques to extract part based features. Finally, we aggregate these four types of deep features via the concatenation operator. We demonstrate the effectiveness of our hybrid deep features on three commonly used scene datasets (MIT-67, Scene-15, and Event-8), in terms of the scene image classification task. Extensive comparisons show that our introduced features can produce state-of-the-art classification accuracies which are more consistent and stable than the results of existing features across all datasets.

\end{abstract}
\begin{IEEEkeywords}
Deep learning, Feature extraction, Hybrid deep features, Image classification, Image representation, Machine learning.
\end{IEEEkeywords}

\section{Introduction}
With the fast development of camera technologies, image classification has been a fundamental problem in image processing. Solving it can benefit a variety of areas relying on images and videos, such as robotics, surveillance, forecasting, and so on. Image features are the mathematical representation of images. 
In general, there are three types of scene images features based on the sources of feature extraction. They are conventional computer vision based features \cite{zeglazi_sift_2016,oliva2005gist,oliva_modeling_2001,dalal2005histograms,wu_centrist:_2011,xiao_mcentrist:_2014,margolin2014otc}, tag-based features \cite{zhang2017image,wang2019task,sitaula2019tag}, and deep learning based features \cite{zhang2017image,gong_multi-scale_2014,guo2016bag,8085139,tang_g-ms2f:_2017,kuzborskij2016naive,zhou2016places,he2016deep,bai2019coordinate}. 
Conventional computer vision based methods \cite{zeglazi_sift_2016,oliva2005gist,oliva_modeling_2001,dalal2005histograms,wu_centrist:_2011,xiao_mcentrist:_2014,margolin2014otc} extract features based on the basic components of images such as texture, color, intensity, gradient, etc. They mainly focus on low-level features and lack details about the context 
in the images (e.g., objects and their spatial relationships). They are not suitable for complex images such as scene images that have intra-class dissimilarities and inter-class similarities. They work well with texture images.

Recent works \cite{zhang2017image,wang2019task,sitaula2019tag} have used annotations of similar images available on the internet to extract tag-based features. Given an image, they first search similar images in the web and extract tag-based features from the descriptions of those similar images. These features are based on the contextual information of images. They did not use the content of images directly.

\begin{figure*}[tb]
    \centering
    \includegraphics[width=0.80\textwidth, height=15cm,keepaspectratio]{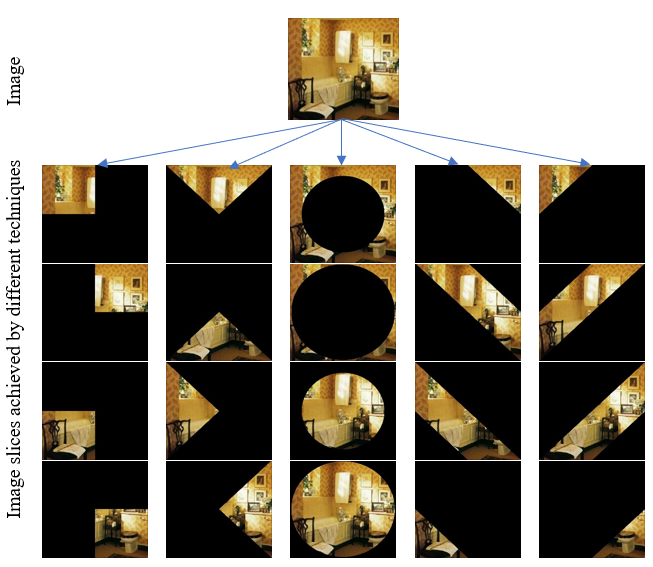}
    \caption{Slices (or sub-images) achieved by using rectangular, triangular, circular, left diagonal cropping and right diagonal cropping techniques, respectively. (columns from left to right order.) We aggregate features of those slices to extract part image level features. Note that the two diagonal slicing techniques are introduced in this work.
    }
    \label{fig:1}
\end{figure*}

More recently, deep features \cite{zhang2017image,gong_multi-scale_2014,guo2016bag,8085139,tang_g-ms2f:_2017,kuzborskij2016naive,zhou2016places,he2016deep,bai2019coordinate} extracted using pre-trained deep learning models have been widely used. They have been shown to work well in various image processing tasks including scene image classification, as they capture high-level semantic information of images. They used deep learning models such as VGG \cite{simonyan2014deep} pre-trained on datasets such as ImageNet \cite{deng_imagenet:_2009} or Places \cite{zhou2016places} to extract features of objects or their background scenes. These techniques employed either object-based (foreground) features or scene-based (background) features. Still, these methods suffer from two problems on scene images. Firstly, existing approaches used either object-based or scene-based features only. For scene images, both object-based and scene-based features should be equally important. 
Secondly, most of these models are pre-trained on images having single objects such as ImageNet \cite{deng_imagenet:_2009}.
But many scene images contain multiples objects and discriminating regions. 
They may not be able to identify some interesting semantic regions in scene images. Some researchers adopted slicing techniques to partition images into smaller parts, and thus part image level features are extracted from image slices \cite{ali2018hybrid}. Whole image level features are also necessary for those scene images containing single objects or other discriminating information like background. Thus, both levels of images are important in extracting the scene features.

In this paper, we assume that features from whole image level and part image level are both useful to identify different interesting regions in scene images, and propose to fuse four types of features -- object-based and scene-based features from both whole images and part images, to construct our hybrid deep features (abbreviated as $HDF$). To get sub-images from an image, triangular, circular and rectangular slicing techniques have been presented in the literature \cite{ali2018hybrid}. To capture more interesting features from part images, we introduce two additional slicing techniques - left and right diagonal slicing \textcolor{blue}{totaling 20 sub-images.
We visually observe and speculate that such 20 sub-images for each scene image can provide optimal semantic regions.}
For example, in Fig. \ref{fig:1}, the image slice at the bottom right corner can identify the object more easily compared to slices in the first three slices of the same row. The aggregation of deep features extracted from sub-images produced by five slicing techniques will construct deep features at part image level, which will complement features from the whole images level. We present some common aggregation operators and empirically select the concatenation operator due to its outstanding aggregation ability. 

Extensive experiments in scene image classification on three commonly used benchmark datasets (MIT-67\cite{quattoni_recognizing_2009}, Scene-15\cite{fei-fei_bayesian_2005} and Event-8\cite{li2007and}) validate the proposed hybrid deep features ($HDF$), and reveal that our $HDF$ generate state-of-the-art classification accuracies which are more consistent and stable than the results of existing features across different datasets. 

The remainder of this paper is organized as follows. Section \ref{related_words}
reviews related works in scene image representation using content and context features. Section \ref{proposed_method} explains our proposed method to extract hybrid deep features ($HDF$). Experimental results and their analysis are discussed in Section \ref{experiment}, which follows Section \ref{conclusion}  for the conclusion of our method with future works.

\section{Related works}
\label{related_words}
In this section, we review the state-of-the-art image feature extraction methods. Depending on the source where the features are extracted from, we generally categorize them into three groups: conventional computer vision methods \cite{zeglazi_sift_2016,oliva2005gist,oliva_modeling_2001,dalal2005histograms,wu_centrist:_2011,xiao_mcentrist:_2014,margolin2014otc,quattoni_recognizing_2009,zhu_large_2010,li2010object,parizi2012reconfigurable,juneja2013blocks,lin_learning_2014,shenghuagao2010local,perronnin2010improving}, tag-based methods \cite{zhang2017image,wang2019task,sitaula2019tag}, and deep learning based methods \cite{zhang2017image,gong_multi-scale_2014,guo2016bag,8085139,tang_g-ms2f:_2017,kuzborskij2016naive,zhou2016places,he2016deep,bai2019coordinate}.

\subsection{Conventional computer vision based methods}
Conventional vision based methods basically rely on the hand-crafted feature extraction techniques such as Generalized
Search Trees (GIST) \cite{oliva2005gist},
GIST-color \cite{oliva_modeling_2001},
Scale-invariant Feature Transform 
(SIFT) \cite{zeglazi_sift_2016},
Histogram of Gradient (HOG) \cite{dalal2005histograms},
Spatial Pyramid Matching (SPM) \cite{lazebnik2006beyond},
CENsus TRansform hISTogram (CENTRIST) \cite{wu_centrist:_2011},
Oriented Texture Curves (OTC) \cite{margolin2014otc},
multi-channel CENTRIST (mCENTRIST) \cite{xiao_mcentrist:_2014},
RoI (regions of interest) with GIST\cite{quattoni_recognizing_2009},
MM (Max-Margin)-Scene\cite{zhu_large_2010},
Object bank\cite{li2010object},
Reconfigurable Bag of Words (RBoW)\cite{parizi2012reconfigurable},
Bag of Parts (BoP)\cite{juneja2013blocks},
Important Spatial Pooling Region (ISPR)\cite{lin_learning_2014},
Laplacian Sparse coding SPM (LscSPM)\cite{shenghuagao2010local},
Improved Fisher Vector (IFV)\cite{perronnin2010improving}
, and so on. All of these features are extracted using the fundamental information of the images such as intensity, colors, orientations, etc. Furthermore, these features are basically relied on the local details, and therefore suitable for certain images such as texture images. They are usually poor for complex images such as scenes. Also, the size of these types of features are often higher than other high-level semantic features.

\subsection{Tag-based methods}
These features are extracted based on the contextual information of images. They represent scene images by tags extracted from annotations/descriptions of similar images available on the web \cite{zhang2017image,wang2019task,sitaula2019tag}. Zhang et al. \cite{zhang2017image} used descriptions of similar images to design bag-of-words (BoW) features directly. In this approach,
there is not only the chance of having outlier tags but also high-dimensional features. To overcome this limitation, Wang et al. \cite{wang2019task} proposed the concept of filter bank using pre-defined categories obtained from ImageNet \cite{deng_imagenet:_2009} and Places \cite{zhou2016places} to filter out the outliers to some extent. Because the filter bank is solely dependent on pre-defined category names, it is more likely to miss other important tags related to images. 
Recently, Sitaula et al. \cite{sitaula2019tag} designed a novel filter bank to extract the tag-based features by exploiting the semantic similarity of tags with image category labels. It provides rich tag-based features and produces better classification accuracies compared to other tag-based features.

\subsection{Deep learning based methods}
In most cases, features extracted from deep learning models \cite{zhang2017image,gong_multi-scale_2014,guo2016bag,8085139,tang_g-ms2f:_2017,kuzborskij2016naive,zhou2016places,he2016deep,bai2019coordinate} 
are found to have more promising classification accuracies for scene images than other methods.
The popular deep learning based feature extraction methods for scene images are:
CNN-MOP\cite{gong_multi-scale_2014},
CNN-sNBNL\cite{kuzborskij2016naive},
VGG\cite{zhou2016places},
ResNet152\cite{he2016deep}
EISR \cite{zhang2017image},
G-MS2F\cite{tang_g-ms2f:_2017},
SBoSP-fusion\cite{guo2016bag},
BoSP-Pre\_gp\cite{8085139},
CNN-LSTM\cite{bai2019coordinate}, and so on.

\begin{figure*}[t]
    \centering
    \includegraphics[width=0.70\textwidth, height=15cm,keepaspectratio]{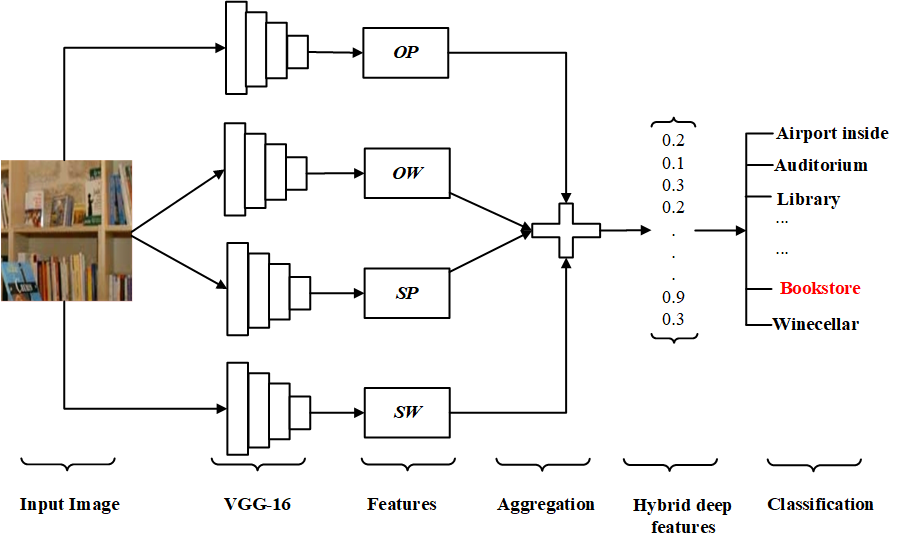}
    \caption{Overview of our approach. The notations $OP$, $OW$, $SP$, and $SW$ represent object-based features at part image level, object-based features at whole image level, scene-based features at part image level, and scene-based features at whole image level, respectively.}
    \label{fig:2}
\end{figure*}

Gong et al. \cite{gong_multi-scale_2014} and Kuzborskij et al. \cite{kuzborskij2016naive} used the Caffe model \cite{jia2014caffe} to extract the multi-scale deep features.
Zhou et al. \cite{zhou2016places} launched a new scene related dataset and trained deep learning architectures such as VGG model \cite{simonyan2014deep}. The features extracted by their method produced promising classification accuracies on scene images.
He et al. \cite{he2016deep} proposed a novel deep learning architecture based on residual concepts and outperforms the previous state-of-the-art deep architectures such as the VGG model \cite{simonyan2014deep}, GoogleNet model \cite{szegedy2014going}, etc.
Zhang et al. \cite{zhang2017image} extracted deep features of an image using its multiple sub-images through random slicing. They concatenated deep features of each slice as a set of deep features of the image. Finally, they combined the deep features with tag-based features to produce a final set of features for the classification purpose.
Tang et al. \cite{tang_g-ms2f:_2017} employed a score-fusion approach to extract deep features. They chose the GoogleNet model \cite{szegedy2014going} and extracted score features from three classification layers for the fusion.
Guo et al. \cite{guo2016bag,8085139} utilized the VGG16 model \cite{simonyan2014deep} to extract unsupervised features by introducing the concept of the bag of surrogate parts (BoSP).
It not only reduced the size of features but also improved the classification accuracies. Furthermore, while comparing different pooling layers of the VGG16 model \cite{simonyan2014deep}, they unveiled that the $5^{th}$ pooling is the best among others in terms of classification accuracy, owing to its better representation capability of objects in the image.
Recently, Bai et al. \cite{bai2019coordinate} designed a new deep model by combining Convolutional Neural Networks (CNNs) with Long Short Term Memory networks (LSTMs). They cast the issue of ordered sliced images as a sequence problem and designed a network to extract scene image features. 

To sum up, the limitations of the existing deep learning based methods are twofold. Firstly, the existing methods extract features at one level only, and ignore a hybrid of features in part image level and whole image level. Aggregating features extracted from both part image level and the whole image can be useful to identify interesting semantic regions in the image. Secondly, the existing methods rely on either scene-based features or object-based features only. Both types of features are equally important for scene images representation. The objects may not be the sole discriminators of the scene images since the contextual information in the image background can change their semantic meanings. 

\section{The Proposed Method}
\label{proposed_method}
We propose to extract hybrid deep features ($HDF$) by fusing scene-based and object-based deep features at both the whole image and part image levels. Our method consists of five steps: object-based features extraction at the part image level, object-based features extraction at the whole image level, scene-based features extraction at the part image level, scene-based features extraction at the whole image level, and aggregation/fusion of the four types of features. 

We employ the VGG16 models \cite{simonyan2014deep} and exploit the $5^{th}$ pooling layer, as suggested by \cite{guo2016bag,8085139}. To extract scene-based and object-based deep features, we use the VGG16 models \cite{simonyan2014deep}  pre-trained on ImageNet \cite{deng_imagenet:_2009} and Places \cite{zhou2016places} datasets. The VGG16 model \cite{simonyan2014deep} pre-trained on ImageNet \cite{deng_imagenet:_2009} provides features related to objects (foreground) in an image, whereas the VGG16 model \cite{simonyan2014deep} pre-trained on Places \cite{zhou2016places} provides features related to the scene (background) in the image. 
We resize all the images into $224\times 224$ before inputting to the VGG16 model.
To extract part level features of an image, we aggregate features extracted from multiple sub-images of the image produced by five different slicing techniques (three existing techniques and two introduced in this work). We perform the Global Average Pooling (GAP) operation on the deep features extracted by the deep learning model to extract the 512-D features. The GAP operation captures both lower and higher activation values in each feature map of the deep learning model, which is suitable for scene images to grab the discriminant features.
\textcolor{blue}{Similarly, motivated by the ability of GAP operation, we apply mean pooling to leverage both higher and lower activation values on the feature vectors extracted by GAP operation.}
The five steps in the proposed method is discussed successively in the next five subsections.

\subsection{Object-based features extraction at the part image level}
To extract object-based parts level ($OP$) features, we slice an image ($I$) into $20$  slices (each of the five slicing techniques yields four image parts, shown in Fig. \ref{fig:1}) \{$I_1, I_2, I_3, \cdots, I_{20}$\}. For each image slice $I_i$, deep features are extracted from the VGG16 model pre-trained on ImageNet using the GAP operation. To obtain $OP$ features of $I$, we then aggregate deep features of the $20$ slices by performing the mean pooling operation as:  
\begin{equation}
    OP(I)=Mean\{OF(I_1),OF(I_2), \cdots, OF(I_{20})\},
    \label{eq:1}
\end{equation}
where $OF(I_i)=GAP\{VGG16_{ImageNet}(I_i)\}$ indicates the GAP operation based deep features of image slice $I_i$ from the $5^{th}$ pooling layer of VGG16 model pre-trained on ImageNet.

\subsection{Object-based features extraction at the whole image level}
To extract the object-based whole image level ($OW$) features, we again adopt the VGG16 model \cite{simonyan2014deep} pre-trained on ImageNet ($VGG16_{ImageNet}$). We extract deep features of the whole image $I$ via the GAP operation from the $5^{th}$ pooling layer of such VGG16 model.
\begin{equation}
    OW(I)=GAP\{VGG16_{ImageNet}(I)\}
    \label{eq:3}
\end{equation}

\subsection{Scene-based features extraction at the part image level}
To extract scene-based parts level ($SP$) features of the image $I$, the deep features of the $20$ slices \{$I_1, I_2, I_3, \cdots, I_{20}$\} extracted from the the $5^{th}$ pooling layer of the VGG16 model pre-trained on Places ($VGG16_{Places}$) are combined through the mean pooling. 
\begin{equation}
    SP(I)=Mean\{SF(I_1),SF(I_2), \cdots, SF(I_{20})\},
    \label{eq:1}
\end{equation}
where $SF(I_i)=GAP\{VGG16_{Places}(I_i)\}$ denotes the GAP based deep features of the image slice $I_i$ from the $5^{th}$ pooling layer of the VGG16 model pre-trained on Places.

\subsection{Scene-based features extraction at the whole image level}
Similarly, $VGG16_{Places}$ is used to extract scene-based whole image level ($SW$) features of image $I$ as: 
\begin{equation}
    SW(I)=GAP\{VGG16_{Places}(I)\}
    \label{eq:6}
\end{equation}

\subsection{Features aggregation}
After computing the above four types of deep features, we need to aggregate them to form the hybrid deep features ($HDF$) for the representation of the scene image $I$. Such features is achieved by using a pooling operator:
\begin{equation}
    HDF(I)=Pool\{OP(I), OW(I), SP(I), SW(I)\}
    \label{eq:7}
\end{equation}

\begin{figure*}[tb]
\begin{center}
 \includegraphics[width=0.90\textwidth, height=50mm,keepaspectratio]{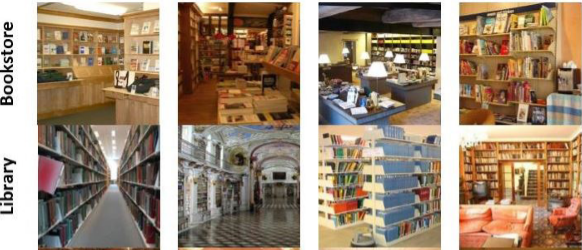}
  \caption{Images sampled from MIT-67 \cite{quattoni_recognizing_2009}. }
  \label{fig:mit67}
 \end{center}
  \end{figure*}
\begin{figure*}[tb]
\begin{center}
 \includegraphics[width=0.90\textwidth, height=50mm,keepaspectratio]{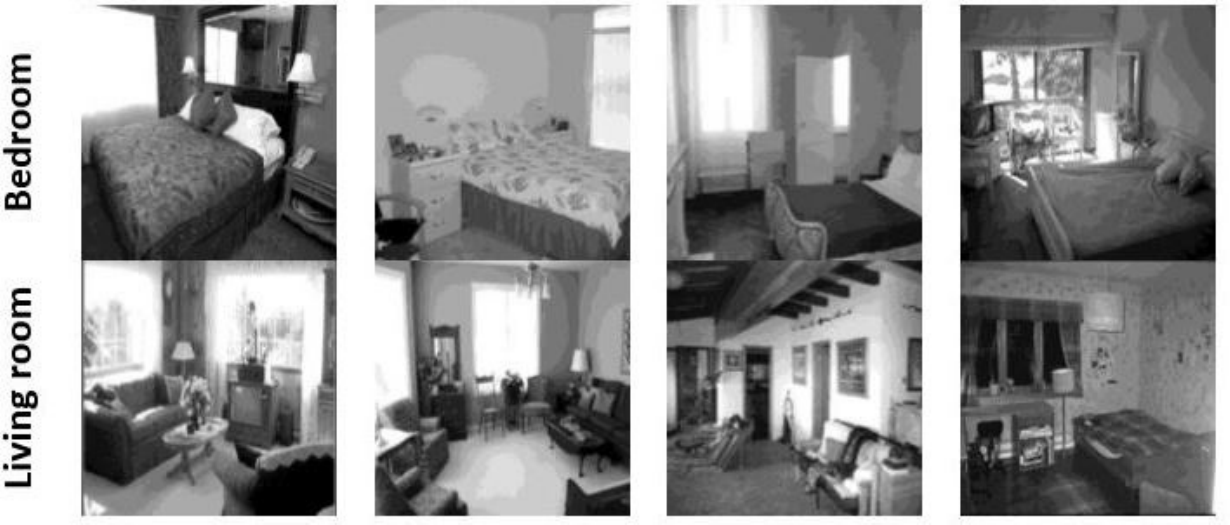}
  \caption{Images sampled from Scene-15 \cite{fei-fei_bayesian_2005}. }
  \label{fig:scene15}
 \end{center}
  \end{figure*}
 \begin{figure*}[tb]
\begin{center}
 \includegraphics[width=0.90\textwidth, height=50mm,keepaspectratio]{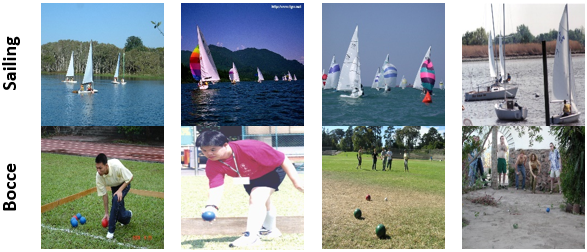}
  \caption{Images sampled from Event-8 \cite{li2007and}. }
  \label{fig:event8}
 \end{center}
  \end{figure*}
Note that the size of each type of deep features is $512$. The size of the final hybrid features depends on the pooling operation. It remains $512$ if Min, Max or Mean pooling is used, but it increases to $2048$ for the concatenation operation. We empirically found that the concatenation produces better results than others. Therefore, all experiments conducted in this work are based on the concatenation operation which leads to $2048$-D features. We present the comparison results of all four pooing operations in Section \ref{sec_ablExp_pooling}.  

Finally, we utilize such hybrid features ($HDF$) obtained from Eq. \eqref{eq:7} for the task of scene image classification. The overview of the proposed method is shown in Fig. \ref{fig:2}.

\section{Experiments and Analysis}
\label{experiment}
In this section, we will explain the used datasets, implementation and experimental results (comparisons and ablation studies) in scene classification.

\subsection{Datasets}
We employ three commonly used scene image datasets: MIT-67\cite{quattoni_recognizing_2009} (Fig. \ref{fig:mit67}), Scene-15\cite{fei-fei_bayesian_2005} (Fig. \ref{fig:scene15}), and Event-8\cite{li2007and} (Fig. \ref{fig:event8}). 

{\bf MIT-67} \cite{quattoni_recognizing_2009} is the largest indoor scene dataset employed in the experiment, and has been used by previous studies \cite{quattoni_recognizing_2009,zhu_large_2010,parizi2012reconfigurable,juneja2013blocks,margolin2014otc,perronnin2010improving,lin_learning_2014,zhang2017image,gong_multi-scale_2014,guo2016bag,8085139,tang_g-ms2f:_2017,kim2014convolutional,wang2019task,sitaula2019tag,bai2019coordinate,he2016deep,li2010object}. It contains $15,620$ images belonging to $67$ indoor categories. We use the same train/test split as suggested by \cite{quattoni_recognizing_2009}. For train/test split, $80$ images per category are used for training and the remaining $20$ images are used for testing.

{\bf Scene-15} \cite{fei-fei_bayesian_2005} dataset comprises images of $15$ categories. There are $4,485$ images, where each category contains $200$ to $400$ images.
We create $10$ train/test sets and present the average accuracy, as done by previous studies \cite{oliva_modeling_2001,lazebnik2006beyond,wu_centrist:_2011,margolin2014otc,lin_learning_2014,perronnin2010improving,zhang2017image,tang_g-ms2f:_2017,kim2014convolutional,wang2019task,he2016deep,sitaula2019tag}. For each train/test set, $100$ images per category are selected for the training set and the rest for the test set.

{\bf Event-8} \cite{li2007and} dataset includes images of $8$ different sports categories.  There are $1,579$ images in this dataset, where each category comprises $137$ to $250$ images.
We also use 10 different train/test sets, as with previous studies \cite{li2010object,lin_learning_2014,shenghuagao2010local,perronnin2010improving,zhang2017image,kuzborskij2016naive,zhou2016places,he2016deep,wang2019task,kim2014convolutional,sitaula2019tag}, and present the average accuracy. For each train/test set, $70$ images per category are involved in the train set and $60$ images are involved in the test set. 

\subsection{Implementation}
To implement our approach, we use the Keras python package \cite{chollet2015keras} for the deep learning models pre-trained on the Places \cite{zhou2016places,gkallia2017keras_places365} and ImageNet \cite{deng_imagenet:_2009} datasets. The proposed hybrid features ($HDF$) are encoded and normalized as suggested by Guo et al. \cite{guo2016bag,8085139}. 
We use the $L_2$-Regularized Logistic Regression classifier (LR) implemented using LibLinear \cite{fan2008liblinear} as the classifier to classify scene images.
\textcolor{blue}{We use such classifier in our experiment because of two reasons: first, it is faster owing to its simple operations and second, it produces better classification accuracy on deep features like ours as suggested by \cite{guo2016bag,8085139}}.
In each experiment, the cost parameter (C) is automatically tuned
in the range \{$1,2,3,\cdots,100$\} using a grid search technique, and the default settings of other parameters are used. All experiments are conducted on a laptop with an NVIDIA GeForce GTX 1050 GPU.
\begin{table}[t] 
\caption{Classification accuracy (\%) of the state-of-the-art methods and our proposed method \textcolor{blue}{on testing set of three datasets}. Best accuracy is in bold and the second best accuracy is underlined.
The asterisk (*) symbol represents no published results on the corresponding dataset. }
\centering
\begin{tabular}{p{2.9cm} p{0.9cm} p{1.2cm} p{0.9cm}}
\toprule
Method & MIT-67 & Scene-15   & Event-8  \\
\midrule
 \multicolumn{4}{l}{Conventional computer vision based methods}\\
\midrule
GIST-color\cite{oliva_modeling_2001} & *&69.5 &* \\
ROI with GIST\cite{quattoni_recognizing_2009} &26.1 &* &*\\
SPM\cite{lazebnik2006beyond} &* &81.4 &* \\
MM-Scene\cite{zhu_large_2010} & 28.3 &* &*\\
CENTRIST\cite{wu_centrist:_2011}&* &83.9 &* \\
Object Bank\cite{li2010object}&37.6 &* &76.3\\
RBoW\cite{parizi2012reconfigurable}&37.9 &* &*\\
BOP\cite{juneja2013blocks}&46.1&* &*\\
OTC\cite{margolin2014otc}&47.3&84.4 &*\\
ISPR\cite{lin_learning_2014}&50.1&85.1 &74.9\\
LscSPM\cite{shenghuagao2010local}&* &* &85.3\\
IFV\cite{perronnin2010improving}&60.8 &89.2 &90.3\\
\midrule
\multicolumn{4}{l}{Tag-based methods}\\
\midrule
BoW\cite{wang2019task} &52.5&70.1 & 83.5 \\
CNN \cite{kim2014convolutional} & 52.0&72.2 &85.9\\
s-CNN(max)\cite{wang2019task} & 54.6&76.2 &90.9 \\
s-CNN(avg)\cite{wang2019task} & 55.1&76.7 &91.2\\
s-CNNC(max)\cite{wang2019task} & 55.9&77.2 &91.5\\
TSF \cite{sitaula2019tag}& 76.5 & 81.3 &94.4 \\
\midrule
\multicolumn{4}{l}{Deep learning-based methods}\\
\midrule
EISR \cite{zhang2017image} &66.2 &\textbf{94.5} & 92.7 \\
CNN-MOP\cite{gong_multi-scale_2014} & 68.0&* &* \\
SBoSP-fusion\cite{guo2016bag} &77.9 &* &* \\
BoSP-Pre\_gp\cite{8085139} & 78.2&* &* \\
G-MS2F\cite{tang_g-ms2f:_2017}& 79.6&92.9 &*\\
CNN-sNBNL\cite{kuzborskij2016naive} &* &* &95.3 \\
VGG\cite{zhou2016places} &* &* &95.6 \\
ResNet152\cite{he2016deep} &77.4 &92.4 &\textbf{96.9}  \\
CNN-LSTM\cite{bai2019coordinate}&\underline{80.5} &*&* \\
\midrule
\textbf{Ours \emph{HDF}}& \textbf{82.0} & \underline{93.9} &\underline{96.2} \\
\hline
\end{tabular}
\label{tab:1}
\end{table}
\subsection{Comparison with the state-of-the-art methods}
We evaluate the classification performance of the proposed hybrid deep features ($HDF$) against the features by $27$ state-of-the-art image features extraction methods ($12$ conventional computer vision based features, $6$ tag-based features and $9$ deep learning based features). The classification accuracies of the proposed features and other contenders are provided in Table \ref{tab:1}. Notice that results of the contenders are taken from the corresponding published papers. 

The results presented in the first column of Table \ref{tab:1} show that the proposed hybrid deep features ($HDF$) produce the best result with an accuracy of $82.0\%$ on the MIT-67 dataset. CNN-LSTM\cite{bai2019coordinate} ($80.5\%$) ranks the second best, followed by G-MS2F\cite{tang_g-ms2f:_2017} ($79.6\%$) in the third place. Results of tag-based features are at least $5.5\%$ worse than $HDF$, and conventional computer vision based methods generate worse results by at least $21.2\%$ than our $HDF$. 

On Scene-15 (results in the second column of Table \ref{tab:1}), the proposed hybrid deep features produce the second best accuracy ($93.9\%$) which is behind $94.5\%$ of EISR \cite{zhang2017image}. G-MS2F\cite{tang_g-ms2f:_2017} ranks the third place with an accuracy of $92.9\%$. Similar to MIT-67, conventional computer vision based features and tag-based features induce worse results than deep learning based features on Scene-15. 

The results in the third column of Table \ref{tab:1} show that the proposed hybrid deep features generate the second best accuracy of $96.2\%$ which is slightly behind $96.9\%$ of ResNet152\cite{he2016deep}. VGG\cite{zhou2016places} achieved the third best result with an accuracy of $95.6\%$. 

Classification results in three datasets show that the proposed hybrid deep features ($HDF$) can produce the best or second best results in all cases. It demonstrates the consistent and stable performance of our hybrid deep features across different datasets. Results of other features vary significantly across different datasets (MIT-67, Scene-15, and Event-8)
. EISR \cite{zhang2017image} has the best result in Scene-15 but ranks the eighth and fifth in the MIT-67 and Event-8 datasets, respectively. Similarly, ResNet152\cite{he2016deep} produces the best result in Event-8, but it ranks the sixth in MIT-67 and the fourth in Scene-15, respectively. 
Taking a closer look at  EISR \cite{zhang2017image}, its features size is extremely higher than ours. They also use random cropping of the image regions which ranges from 40 to 400 and concatenate the deep features of all the regions sequentially, and then they concatenate the features with tag-based features. As a result, they have far more than $81,920$-D features, with considering the concatenation with the tag-based features. This is probably why it generates the best accuracy on Scene-15.  
Similarly, ResNet152 \cite{he2016deep} provides $2,048$-D features, the size of which is equal to ours. However, it induces the highest accuracy in only one dataset and much worse accuracies in the other two datasets, while our features are more stable and perform the best or the second best across all three datasets.
We suspect that the consistent and stable performance of the proposed hybrid deep features across three datasets is mainly due to the fusion of object-based and scene-based features at both whole image and parts levels. The four types of features enable the capture of different and complementary information on one image.

\subsection{Ablative study of individual features }
We also evaluate the performances of each of the four types of features. The classification accuracies using object-based features on part images ($OP$), object-based features on the whole image ($OW$), scene-based features on part images ($SP$), and scene-based features on the whole image ($SW$) are provided in Table \ref{tab:2}. 

By observing results in Table \ref{tab:2}, we notice that the $SW$ features resulted in the best accuracy on the MIT-67 and Scene-15 datasets ($79.7\%$ and $92.8\%$, respectively), whereas the $OW$ features induce the best result on the Event-8 dataset ($95.7\%$). The images on Event-8 often contain single objects which can be easily captured by the object-based features at the whole image level ($OW$). However, images on MIT-67 and Scene-15 are more dependent on scenes, and thus scene-based features $SW$ bring about the best accuracy.
These three best accuracies are obviously lower than the accuracies led by the hybrid features of all four types of features, which further demonstrates the necessity of the aggregation of the four types of features. 
\begin{table}[t] 
\caption{Classification accuracy (\%) of each individual type of features ($OP$, $OW$, $SP$, and $SW$) \textcolor{blue}{on testing set of three datasets}.
}
\centering
\begin{tabular}{p{1.2cm} p{0.9cm} p{0.9cm} p{0.9cm} p{0.9cm}} 
\toprule
Dataset & $OP$ & $OW$ & $SP$ & $SW$ \\ 
\midrule
MIT-67  &68.2 &70.7 &76.5 &\textbf{79.7}\\ 
Scene-15  &88.8 &90.3 &92.1 &\textbf{92.8}\\ 
Event-8  &93.7 &\textbf{95.7} &93.0 &94.9\\ 
\hline
\end{tabular}
\label{tab:2}
\end{table}
\begin{table}[t] 
\caption{Classification accuracy (\%) of our hybrid deep features ($HDF$) achieved by four different aggregation methods (Max, Mean, Min, and Concatenate) \textcolor{blue}{on testing set of three datasets}. }
\centering
\begin{tabular}{p{1.2cm} p{0.9cm} p{0.9cm} p{0.9cm} p{0.9cm}}
\toprule
Dataset & Max& Mean& Min & Concat\\
\midrule
MIT-67  &79.9 &80.3 &67.9 &\textbf{82.0}\\
Scene-15 &93.3 &93.4 &87.4 &\textbf{93.9} \\
Event-8  &96.0 &95.8 &90.9  &\textbf{96.2}\\
\hline
\end{tabular}
\label{tab:4}
\end{table}

\subsection{Ablative study of aggregation methods}
\label{sec_ablExp_pooling}
To study the efficacy of the four aggregation methods (Max pooling, Mean pooling, Min pooling, and Concat pooling in Eq. \eqref{eq:7},   
we perform experiments on all three datasets, and the results are summarized in Table \ref{tab:4}. The results manifests that our proposed features ($HDF$) by the Concat aggregation yield higher accuracies than other methods on all three datasets. This is because all these features are different types of features capturing different information about images.

\section{Conclusion}
\label{conclusion}
In this paper, we have introduced hybrid deep features to represent images by aggregating four types of features (scene-based and object-based features at the whole image and part image levels). Since the four types of deep features capture different types of information about images, fusing them together can provide richer discriminant information of images. Experimental results in scene image classification on three widely used scene image datasets unveil that the proposed hybrid deep features are capable of producing more consistent and stable results (the best or second best) than the state-of-the-art techniques. We also notice that the proposed features are more prominent and suitable for indoor scene images because such images contain both objects and scenes as the discriminator information.

Compared to indoor scene images, the proposed features may be less powerful in representing other types of images like outdoor images. 
Furthermore, we only used the $5^{th}$ pooling layer in our method, which may not be sufficient to extract features. In the future, we would like to analyze the characteristics of different types of images and exploit other layers of the pre-trained models for the classification task.

\bibliographystyle{IEEEtran}
\bibliography{references}

\begin{thebibliography}{10}
\providecommand{\url}[1]{#1}
\csname url@samestyle\endcsname
\providecommand{\newblock}{\relax}
\providecommand{\bibinfo}[2]{#2}
\providecommand{\BIBentrySTDinterwordspacing}{\spaceskip=0pt\relax}
\providecommand{\BIBentryALTinterwordstretchfactor}{4}
\providecommand{\BIBentryALTinterwordspacing}{\spaceskip=\fontdimen2\font plus
\BIBentryALTinterwordstretchfactor\fontdimen3\font minus
  \fontdimen4\font\relax}
\providecommand{\BIBforeignlanguage}[2]{{%
\expandafter\ifx\csname l@#1\endcsname\relax
\typeout{** WARNING: IEEEtran.bst: No hyphenation pattern has been}%
\typeout{** loaded for the language `#1'. Using the pattern for}%
\typeout{** the default language instead.}%
\else
\language=\csname l@#1\endcsname
\fi
#2}}
\providecommand{\BIBdecl}{\relax}
\BIBdecl

\bibitem{zeglazi_sift_2016}
O.~Zeglazi, A.~Amine, and M.~Rziza, ``Sift {descriptors} {modeling} and
  {application} in {texture} {image} {classification},'' in \emph{Proc. 13th
  Int. Conf. Comput. Graphics, Imaging and Visualization (CGiV)}, Mar. 2016,
  pp. 265--268.

\bibitem{oliva2005gist}
A.~Oliva, ``Gist of the scene,'' in \emph{Neurobiology of Attention}.\hskip 1em
  plus 0.5em minus 0.4em\relax Elsevier, 2005, pp. 251--256.

\bibitem{oliva_modeling_2001}
A.~Oliva and A.~Torralba, ``Modeling the {shape} of the {scene}: {a} {holistic}
  {representation} of the {spatial} {envelope},'' \emph{Int. J. Comput. Vis.},
  vol.~42, no.~3, pp. 145--175, May. 2001.

\bibitem{dalal2005histograms}
N.~Dalal and B.~Triggs, ``Histograms of oriented gradients for human
  detection,'' in \emph{Proc. IEEE Comput. Soc. Conf. Comput. Vis. Pattern
  Recognit. (CVPR)}, 2005, pp. 886--893.

\bibitem{wu_centrist:_2011}
J.~Wu and J.~M. Rehg, ``{CENTRIST}: {A} {visual} {descriptor} for {scene}
  {categorization},'' \emph{{IEEE} Trans. Pattern Anal. Mach. Intell.},
  vol.~33, no.~8, pp. 1489--1501, Aug. 2011.

\bibitem{xiao_mcentrist:_2014}
Y.~Xiao, J.~Wu, and J.~Yuan, ``{mCENTRIST}: {a} {multi}-{channel} {feature}
  {generation} {mechanism} for {scene} {categorization},'' \emph{{IEEE} Trans.
  Image Process.}, vol.~23, no.~2, pp. 823--836, Feb. 2014.

\bibitem{margolin2014otc}
R.~Margolin, L.~Zelnik-Manor, and A.~Tal, ``Otc: A novel local descriptor for
  scene classification,'' in \emph{Proc. Eur. Conf. Comput. Vis. (ECCV)}, 2014,
  pp. 377--391.

\bibitem{zhang2017image}
C.~Zhang, G.~Zhu, Q.~Huang, and Q.~Tian, ``Image classification by search with
  explicitly and implicitly semantic representations,'' \emph{Information
  Sciences}, vol. 376, pp. 125--135, 2017.

\bibitem{wang2019task}
D.~Wang and K.~Mao, ``Task-generic semantic convolutional neural network for
  web text-aided image classification,'' \emph{Neurocomputing}, vol. 329, pp.
  103--115, 2019.

\bibitem{sitaula2019tag}
C.~Sitaula, Y.~Xiang, A.~Basnet, S.~Aryal, and X.~Lu, ``Tag-based semantic
  features for scene image classification,'' in \emph{Proc. Int. Conf. on
  Neural Inf. Process. (ICONIP)}, 2019, pp. 90--102.

\bibitem{gong_multi-scale_2014}
Y.~Gong, L.~Wang, R.~Guo, and S.~Lazebnik, ``Multi-scale {orderless} {pooling}
  of {deep} {convolutional} {activation} {features},'' in \emph{Proc. Eur.
  Conf. Comput. Vis. (ECCV)}, 2014, pp. 392--407.

\bibitem{guo2016bag}
Y.~Guo and M.~S. Lew, ``Bag of surrogate parts: one inherent feature of deep
  cnns.'' in \emph{Proc. BMVC}, 2016.

\bibitem{8085139}
Y.~Guo, Y.~Liu, S.~Lao, E.~M. Bakker, L.~Bai, and M.~S. Lew, ``Bag of surrogate
  parts feature for visual recognition,'' \emph{IEEE Trans. Multimedia},
  vol.~20, no.~6, pp. 1525--1536, Jun. 2018.

\bibitem{tang_g-ms2f:_2017}
P.~Tang, H.~Wang, and S.~Kwong, ``G-{MS2F}: {GoogLeNet} based multi-stage
  feature fusion of deep {CNN} for scene recognition,'' \emph{Neurocomputing},
  vol. 225, pp. 188 -- 197, Feb. 2017.

\bibitem{kuzborskij2016naive}
I.~Kuzborskij, F.~Maria~Carlucci, and B.~Caputo, ``When naive bayes nearest
  neighbors meet convolutional neural networks,'' in \emph{Proc. IEEE Conf. on
  Computer Vision and Pattern Recognition (CVPR)}, 2016, pp. 2100--2109.

\bibitem{zhou2016places}
B.~Zhou, A.~Khosla, A.~Lapedriza, A.~Torralba, and A.~Oliva, ``Places: An image
  database for deep scene understanding,'' \emph{arXiv preprint
  arXiv:1610.02055}, 2016.

\bibitem{he2016deep}
K.~He, X.~Zhang, S.~Ren, and J.~Sun, ``Deep residual learning for image
  recognition,'' in \emph{Proc. IEEE Conf. on Computer Vision and Pattern
  Recognition (CVPR)}, 2016, pp. 770--778.

\bibitem{bai2019coordinate}
S.~Bai, H.~Tang, and S.~An, ``Coordinate cnns and lstms to categorize scene
  images with multi-views and multi-levels of abstraction,'' \emph{Expert
  Systems with Applications}, vol. 120, pp. 298--309, 2019.

\bibitem{simonyan2014deep}
K.~Simonyan and A.~Zisserman, ``Very deep convolutional networks for
  large-scale image recognition,'' 2014.

\bibitem{deng_imagenet:_2009}
J.~Deng, W.~Dong, R.~Socher, L.-J. Li, K.~Li, and L.~Fei-Fei, ``{ImageNet}: {a}
  {large}-{scale} {hierarchical} {image} {database},'' in \emph{Proc. {IEEE}
  {Conf.} {Comput.} {Vis.} {Pattern} {Recognit.} ({CVPR})}, 2009.

\bibitem{ali2018hybrid}
N.~Ali, B.~Zafar, F.~Riaz, S.~H. Dar, N.~I. Ratyal, K.~B. B., M.~K. Iqbal, and
  M.~Sajid, ``A hybrid geometric spatial image representation for scene
  classification,'' \emph{PloS one}, vol.~13, no.~9, p. e0203339, 2018.

\bibitem{quattoni_recognizing_2009}
A.~Quattoni and A.~Torralba, ``Recognizing indoor scenes,'' in \emph{Proc.
  {IEEE} {Conf.} {Comput.} {Vis.} {Pattern} {Recognit.} ({CVPR})}, Jun. 2009,
  pp. 413--420.

\bibitem{fei-fei_bayesian_2005}
L.~Fei-Fei and P.~Perona, ``A bayesian hierarchical model for learning natural
  scene categories,'' in \emph{Proc. {IEEE} {Comput.} {Soc.} {Conf.} {Comput.}
  {Vis.} and {Pattern} {Recognit.} ({CVPR})}, vol.~2, Jun. 2005, pp. 524--531.

\bibitem{li2007and}
L.-J. Li and F.-F. Li, ``What, where and who? classifying events by scene and
  object recognition.'' in \emph{ICCV}, vol.~2, no.~5, 2007, p.~6.

\bibitem{zhu_large_2010}
J.~Zhu, L.-j. Li, L.~Fei-Fei, and E.~P. Xing, ``Large {margin} {learning} of
  {upstream} {scene} {understanding} {models},'' in \emph{Proc. Adv. Neural
  Inf. Process. Syst. (NIPS)}, 2010, pp. 2586--2594.

\bibitem{li2010object}
L.-J. Li, H.~Su, L.~Fei-Fei, and E.~P. Xing, ``Object bank: A high-level image
  representation for scene classification \& semantic feature sparsification,''
  in \emph{Proc. Adv. Neural Inf. Process. Syst. (NIPS)}, 2010, pp. 1378--1386.

\bibitem{parizi2012reconfigurable}
N.~Parizi, J.~G. Oberlin, and P.~F. Felzenszwalb, ``Reconfigurable models for
  scene recognition,'' in \emph{Proc. Comput. Vis. Pattern Recognit.(CVPR)},
  Jun. 2012, pp. 2775--2782.

\bibitem{juneja2013blocks}
M.~Juneja, A.~Vedaldi, C.~Jawahar, and A.~Zisserman, ``Blocks that shout:
  Distinctive parts for scene classification,'' in \emph{Proc. IEEE Conf.
  Comput. Vis. Pattern Recognit. (CVPR)}, Jun. 2013, pp. 923--930.

\bibitem{lin_learning_2014}
D.~Lin, C.~Lu, R.~Liao, and J.~Jia, ``Learning {important} {spatial} {pooling}
  {regions} for {scene} {classification},'' in \emph{Proc. IEEE Conf. Comput.
  Vis. Pattern Recognit. (CVPR)}, Jun. 2014, pp. 3726--3733.

\bibitem{shenghuagao2010local}
I.-H. ShenghuaGao and P.~Liang-TienChia, ``Local features are not
  lonely--laplacian sparse coding for image classification,'' pp. 3555--3561,
  2010.

\bibitem{perronnin2010improving}
F.~Perronnin, J.~S{\'a}nchez, and T.~Mensink, ``Improving the fisher kernel for
  large-scale image classification,'' in \emph{Proc. European Conference on
  Computer vision (ECCV)}, 2010, pp. 143--156.

\bibitem{lazebnik2006beyond}
S.~Lazebnik, C.~Schmid, and J.~Ponce, ``Beyond bags of features: Spatial
  pyramid matching for recognizing natural scene categories,'' in \emph{Proc.
  IEEE Comput. Soc. Conf. Comput. Vis. Pattern Recognit.}, Jun. 2006, pp.
  2169--2178.

\bibitem{jia2014caffe}
Y.~Jia, E.~Shelhamer, J.~Donahue, S.~Karayev, J.~Long, R.~Girshick,
  S.~Guadarrama, and T.~Darrell, ``Caffe: Convolutional architecture for fast
  feature embedding,'' in \emph{Proc. 22nd ACM Int. Conf. on Multimedia}, 2014,
  pp. 675--678.

\bibitem{szegedy2014going}
C.~Szegedy, W.~Liu, Y.~Jia, P.~Sermanet, S.~Reed, D.~Anguelov, D.~Erhan,
  V.~Vanhoucke, and A.~Rabinovich, ``Going deeper with convolutions,'' 2014.

\bibitem{kim2014convolutional}
Y.~Kim, ``Convolutional neural networks for sentence classification,''
  \emph{arXiv preprint arXiv:1408.5882}, 2014.

\bibitem{chollet2015keras}
F.~Chollet, ``Keras,'' \url{https://github.com/fchollet/keras}, 2015.

\bibitem{gkallia2017keras_places365}
G.~Kalliatakis, ``Keras-vgg16-places365,''
  \url{https://github.com/GKalliatakis/Keras-VGG16-places365}, 2017.

\bibitem{fan2008liblinear}
R.-E. Fan, K.-W. Chang, C.-J. Hsieh, X.-R. Wang, and C.-J. Lin, ``Liblinear: A
  library for large linear classification,'' \emph{Journal of Machine Learning
  Research}, vol.~9, no. Aug, pp. 1871--1874, 2008.

\end{thebibliography}
\end{document}